# Configuração e Operação da Plataforma Clearpath Husky A200 e Manipulador Cobot UR5 2-finger Gripper

Sodre Hiago, hiago.sodre@utec.edu.uy[1]
Barcelona Sebastian, sebastian.barcelona@utec.edu.uy[1]
Sandin Vincent, vincent.sandin@estudiantes.utec.edu.uy[1]
Moraes Pablo, pablo.moraes@utec.edu.uy[1]
Peters Christopher, qristopherp@gmail.com[2]
da Silva Angél, angel.dasilva@estudiantes.utec.edu.uy[1]
Flores Gabriela, luisa.flores@estudiantes.utec.edu.uy[1]
Mazondo Ahilen, ahilen.mazondo@estudiantes.utec.edu.uy[1]
Fernández Santiago, santiago.fernandez@estudiantes.utec.edu.uy[1]

Assunção Nathalie, nathalie.assuncao@utec.edu.uy[1]
de Vargas Bruna, bruna.devargas@utec.edu.uy[1]
Grando Ricardo, ricardo.bedin@utec.edu.uy[1]
Kelbouscas André, andre.dasilva@utec.edu.uy[1]

[1]Universidad Tecnológica del Uruguay
[2]Ostfalia University of Applied Sciences

**Abstract:** *This article presents initial configuration work and use of the robotic platform and manipulator in question. The development of the ideal configuration for using this robot serves as a guide for new users and also validates its functionality for use in projects. Husky is a large payload capacity and power systems robotics development platform that accommodates a wide variety of payloads, customized to meet research needs. Together with the Cobot UR5 Manipulator attached to its base, it expands the application area of its capacity in projects. Advances in robots and mobile manipulators have revolutionized industries by automating tasks that previously required human intervention. These innovations alone increase productivity but also reduce operating costs, which makes the company more competitive in an evolving global market. Therefore, this article investigates the functionalities of this robot to validate its execution in robotics projects.*

**Resumo:** *Este artigo apresenta um trabalho de configuração inicial e utilização da plataforma robótica e manipulador em questão, o desenvolvimento da configuração ideal para utilização de este robô serve como guia para novos usuários e também valida sua funcionalidade para disposição ao uso em projetos. Husky é uma plataforma de desenvolvimento robótico de grande capacidade de carga útil e sistemas de energia que acomoda uma ampla variedade de cargas úteis, personalizadas para atender às necessidades de pesquisa. Juntamente com o Manipulador Cobot UR5 acoplado a sua base, amplia a área de aplicações de sua capacidade em projetos. Los avances en robots y manipuladores móviles han revolucionado las industrias al automatizar tareas que antes requerían intervención humana. Estas innovaciones no sólo aumentan la productividad sino que también reducen los costos operativos, lo que hace que la empresa sea más competitiva en un mercado global en evolución. Portanto, este artigo investiga as funcionalidades deste robô para a validação de sua execução em projetos da área de robótica.*

Palavras chave: Robótica, Plataforma, Manipulador, Sistema, Investigação.

## 1 - INTRODUÇÃO

O Robô Husky da Clearpath Robotics é um veículo robótico que pode ser equipado com diferentes sensores e ferramentas, o que permite aos pesquisadores explorar uma ampla gama de aplicações e cenários em suas pesquisas. Conhecido por sua robustez e versatilidade, pode ser utilizado em uma variedade de aplicações, como navegação autônoma em ambientes desafiadores e realização de tarefas de transporte. (Clearpath Robotics, 2024). O Manipulador UR5 acoplado ao Husky é um dos maiores robôs colaborativos industriais de carga útil leve. Para aplicações que exigem pequeno porte com alcance e carga útil suficientes, o UR5 ajudará a executar tarefas precisas e meticulosas. (*UR5e Lightweight, Versatile Cobot*, s. f.).

Partindo de uma compra realizada pela Universidade Tecnológica do Uruguai, o robô Husky foi adquirido no início do segundo semestre do ano de 2024, uma compra de grande valor para a instituição que levou um processo de dois anos até sua realização. Devido ao formato de aquisição, foi necessária a realização de uma validação técnica e comprovação da funcionalidade correta de tal robô, assim estudantes do laboratório de robótica e inteligência artificial da universidade, realizaram as devidas avaliações que serão apresentadas neste artigo.

Um robô com uma alta tecnologia e aplicabilidade como este, possibilita uma alta gama de exploração para a área de inteligência artificial e navegação robótica. Alguns projetos desenvolvidos nesta área com o mesmo robô apresentam ótimos resultados de aprendizagem de navegação e detecção de terreno, como exemplo do trabalho realizado de "GA-Nav: Segmentação Eficiente de Terreno para Navegação de Robôs em Ambientes externos não estruturados", onde os autores apresentam uma implementação de seu algoritmo de navegação em um robô Husky para demonstrações de navegação no mundo real, apresentando resultados significativos na implementação real. (*GA-Nav: Efficient Terrain Segmentation For Robot Navigation In Unstructured Outdoor Environments*, 2022).

Investigações como esta, comprovam a alta capacidade de implementação de algoritmos e projetos de investigação a um robô com este. O robô adquirido em questão também conta com um braço robótico de grande versatilidade e um sensor lidar que ampliam ainda mais sua utilização a projetos de robótica, pelos motivos de sua alta configuração de fábrica, este trabalho apresenta suas características técnicas e configurações iniciais de utilização. Além de servir como base teórica para utilização do robô no contexto a aplicar, este trabalho valida suas especificações e funcionalidades com um relatório técnico de configuração inicial no contexto de software e controle, explorando a instalação de dependências e movimentos básicos a realizar com a plataforma robótica.

## 2 - REFERENCIAL TEÓRICO

### 2.1 - CLEARPATH HUSKY A200

Husky é uma plataforma robótica que foi desenhada com uma escalável e aberta arquitetura, o tornando ideal para testes e desenvolvimento de sistemas multi robóticos. Utilizando um protocolo serial de código aberto e oferece suporte de API para ROS e opções para C++ e Python, facilitando a implementação de projetos para integração com pesquisas existentes e a crescente base de conhecimento da próspera comunidade ROS, produzindo resultados de pesquisa com mais rapidez. (Clearpath Robotics, 2024). Uma imagem ilustrativa da plataforma se visualiza na Figura 1.

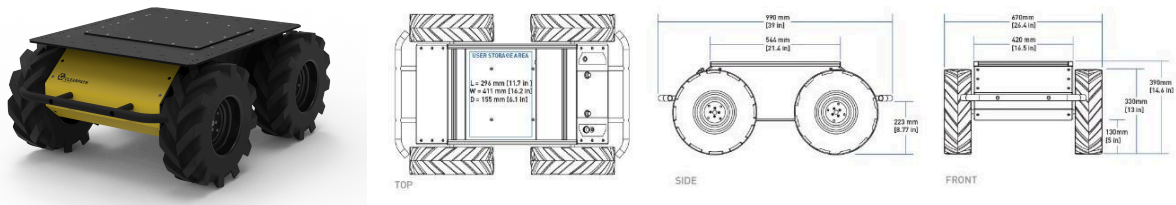

Figura 1. Imagem ilustrativa Clearpath Husky A200. (Clearpath Robotics, 2024).

Sua base robusta e de fácil movimentação permite com que o robô se adapte a ambientes distintos, as rodas de configuração "Off-road", se movimentam sobre uma superfície lisa facilmente e trazem uma melhor tração para ambientes com alta variação de elevação. Além disso conta com um controle bem ajustado, oferecendo ao usuário perfis de movimento incrivelmente suaves mesmo em velocidades lentas (<1cm/s) e com excelente rejeição de perturbações. Aliada às funcionalidades do Husky, ele conta com um design elegante e simples construído com materiais duráveis com poucas peças móveis. Seu sistema de transmissão de alto desempenho e livre de manutenção e pneus com banda de rodagem grande permitem que enfrente terrenos desafiadores do mundo real, prometendo anos de pesquisa produtiva. (Clearpath Robotics, 2024). Na Figura 2 continuação é possível visualizar suas dimensões construtivas.

**Tabela 1. Especificações Técnicas Plataforma**

| | | | |
|---|---|---|---|
| **Dimensões Externas** | 990 x 670 x 390 mm (39 x 26,4 x 14,6 pol.) | **Velocidade Máxima** | 1,0 m/s (2,2 mph) |
| **Dimensões Internas** | 296 x 411 x 155 mm (11,7 x 16,2 x 6,1 pol.) | **Tempo de execução (uso típico)** | 3 horas |
| **Peso** | 50 kg (110 libras) | **Poder de uso** | 5V, 12V e 24V com fusível de 5A cada |
| **Carga útil máxima** | 75 kg (165 libras) | **Drivers e APIs** | ROS Melodic, ROS Kinetic, Biblioteca C++, Mathworks |

Este tipo de plataforma é desenhada para que além da carga de peso suportada, possa ser integrada a tipos de sensores e atuadores ampliando sua utilidade. Juntamente com a empresa que a desenvolve, outras empresas que trabalham com estes dispositivos já realizam a adaptação para implementação na plataforma, facilitando ao usuário. A empresa fornecedora oferece pacotes de produtos com diversos sensores que já podem vir incluídos no seu Kit desenvolvedor. Na Figura 3 a continuação é possível visualizar alguns componentes que podem ser integrados à plataforma e suas desenvolvedoras.

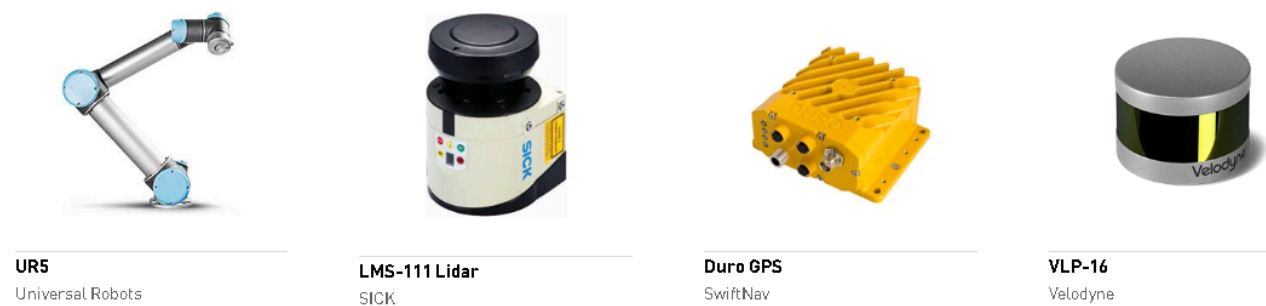

Figura 2. Expansões disponíveis para a plataforma Husky. (Clearpath Robotics, 2024).

## 2.2 - MANIPULADOR COBOT UR5 E 2F-140 FINGER GRIPPER

Desenvolvido pela empresa Universal Robots, que é um fabricante dinamarquês de braços robóticos colaborativos industriais flexíveis menores com sede em Odense na Dinamarca, o manipulador Cobot UR5 segundo sua linha, é o mais complexo dos robôs colaborativos industriais de carga útil leve.

Lançada pela primeira vez em 2008, a versão e-Series de um cobot (robô colaborativo) com carga útil de 5 kg (11 lbs) recebeu melhorias em tecnologia e continua seu caminho para se integrar perfeitamente a uma ampla gama de aplicações. O UR5 é ideal para as necessidades de um cobot leve o suficiente para ser montado em sua linha de produção existente (por exemplo, em uma mesa ou dentro de uma máquina). Correspondendo à necessidade de levantar objetos leves a médios, o UR5e alcança 200 mm mais longe do que o braço de um trabalhador médio (*UR5e Lightweight, Versatile Cobot*, s. f.). É possível visualizar uma imagem ilustrativa deste manipulador na Figura 4 e suas especificações técnicas descritas pela Tabela 2.

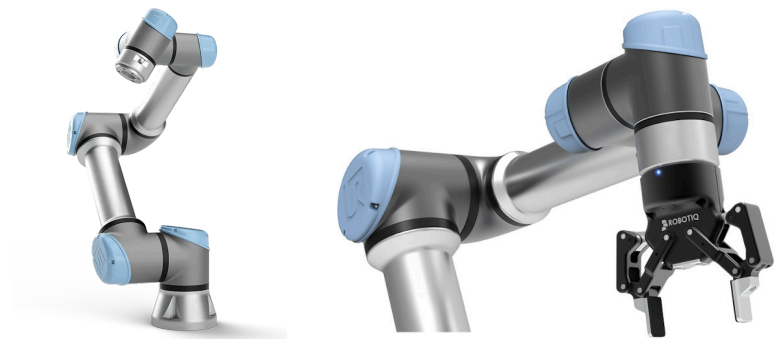

Figura 3. Imagem ilustrativa Cobot UR5 (UR5e Lightweight, Versatile Cobot, s. f.)

**Tabela 2. Especificações Técnicas Manipulador**

| | | | |
|---|---|---|---|
| **Carga útil no deslocamento total** | 5 kg (11 libras) | **Consumo de Energía** | 570 W |
| **Alcance** | 850 mm (33,5 pol) | **Tensão de Alimentação** | 12V / 24V |
| **Graus de liberdade** | 6 articulações rotativas | **Materiais** | Alumínio, Plástico e Aço |
| **Precisão** | 3,5 N (0,2 Nm) | **Peso** | 20,6 kg (141,1 libras) |

Juntamente ao manipulador se encontra a pinça 2F-140 desenvolvida pelo fabricante Robotiq, que amplia a utilização deste manipulador permitindo a manipulação de objetos com seu movimento de agarre, o design patenteado dos dedos permite uma pegada paralela interna e externa, bem como um modo de pegada envolvente exclusivo (2F-85 And 2F-140 Grippers - Robotiq, s. f.). Uma imagem ilustrativa desta pinça é possível visualizar na Figura 3.

## 3 - METODOLOGÍA

### 3.1 - GESTÃO DE RECEBIMENTO DO EQUIPAMENTO

A versão adquirida do robô *Husky A200* pela Universidade Tecnológica - UTEC, acompanha o manipulador *Cobot UR5* com pinça *E 2F-140*, um sensor *Lidar VLP-16* e câmera *Intel RealSense D435i*. Uma imagem do kit completo do robô com as expansões se encontra representada na Figura 4.

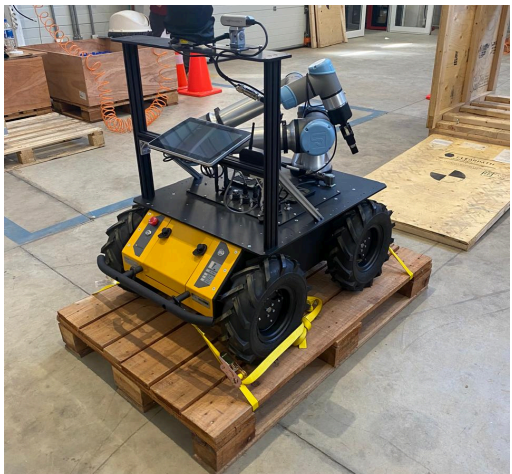
Figura 4. Imagem do robô Husky adquirida no processo de recebimento do equipamento.

No dia 12 de agosto de 2024, o pacote com o robô chegou ao centro logístico da Universidade Tecnológica del Uruguay (UTEC) no polo ITR Norte, em Rivera. Como procedimento padrão de recebimento de importações foi necessário que um docente responsável pelo Laboratório de Robótica e Inteligência Artificial acompanhasse o processo logístico de transferência do equipamento até a sua localização final. Neste caso, o docente responsável além de realizar a verificação do equipamento, registrou cada etapa para o informe técnico do equipamento. Nas Figuras 5 e 6 são apresentados parte do processo logístico comentado e as devidas medidas de seguranças para a manipulação do equipamento em seu processo logístico.

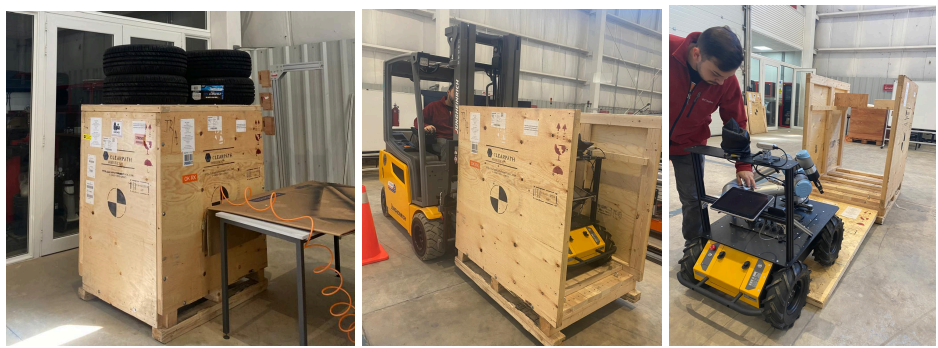
Figura 5. Recepção da caixa do produto.

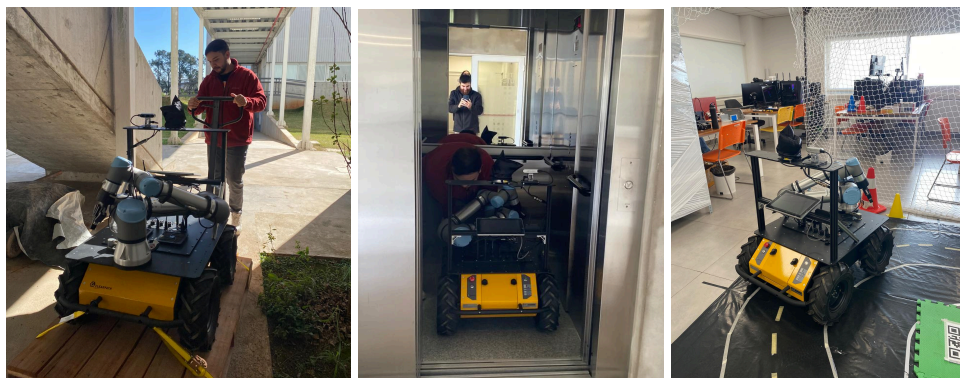
Figura 6. Transporte do ponto de recebimento ao Laboratório de Robótica e IA da UTEC.

**3.2 - CONFIGURAÇÃO INICIAL DO EQUIPAMENTO**

O equipamento acompanha um manual de instruções de configuração em formato físico para o usuário poder configurar o robô e realizar suas devidas conexões. Uma imagem ilustrativa da localização das conexões básicas de baterias e componentes eletrônicos, são representadas na Figura 7. Como procedimento inicial de verificação, foi necessário verificar a integridade das baterias que acompanham o equipamento, com a finalidade de prevenir qualquer descarga ou má conexão ao robô no seu momento de inicialização. Utilizando um multímetro adequado, foi realizada a verificação e funcionalidade da bateria, na Figura 8 se visualiza o procedimento realizado.

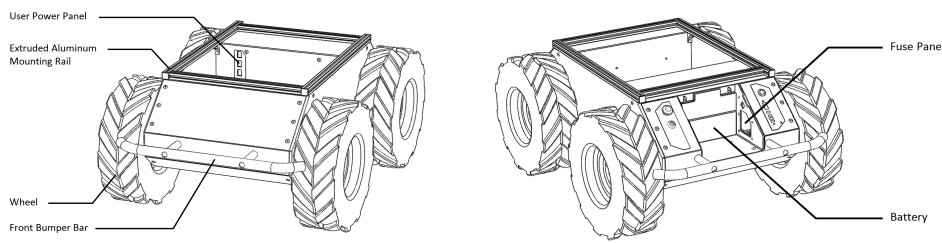

Figura 7. Conexões Husky A200. (*Husky User Manual | Clearpath Robotics Documentation*, 2023).

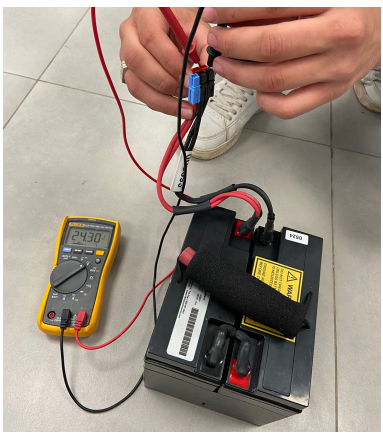

Figura 8. Verificação de integridade de baterias.

Esse robô tem suporte a segunda versão de ROS (sistema operacional de robôs), possibilitando ter acesso a todas suas funcionalidades via sistema operacional e controlá-lo desta forma de maneira mais simplificada. O ROS 2 Humble usa o Ubuntu 22.04. Embora outros sistemas operacionais sejam suportados, optou-se por manter o sistema Ubuntu 22.04 como sistema operacional do robô.

Nesta versão da compra, o robô já vem com a imagem ISO configurada para facilitar a configuração do usuário, porém o ClearPath fornece uma imagem de instalação levemente personalizada do Ubuntu 22.04 que também possui o script necessário para configurar o robô. Para instalar o software em um robô físico por meio da imagem ISO do ClearPath Robotics, é preciso primeiro de uma unidade USB de pelo menos 4 GB para criar a mídia de instalação, um cabo ethernet, um monitor e um teclado, ou também se pode conectar de maneira remota ao robô, configurando uma rede. (*Robot Installation | Clearpath Robotics Documentation*, 2024).

Com a imagem já configurada, seguiu-se os passos do guia de configuração do robô fornecido pela fabricante. (*Robot Installation | Clearpath Robotics Documentation*, 2024).

**3.3 VALIDAÇÃO DE FUNCIONAMIENTO DE COMPONENTES**

Após o processo de configuração dos acessos ao robô, foi possível realizar os testes de funcionamento necessários para a validação técnica das funcionalidades do equipamento. Foram aplicados movimentos de translação e rotação em todos os eixos de movimento da plataforma robótica, para comprovar que não existiam limitações mecânicas em suas rodas ou sensores de aproximação.

Já com o manipulador, sua tela acoplada ao robô permite acesso a interface de movimentação, assim foi possível movimentar em todos eixos cartesianos de translação e rotação do manipulador e também provar a funcionalidade da pinça em pressionar objetos. Por fim, foram feitos testes em sensores e tópicos de posicionamento do robô. Na Figura 8, é possível visualizar o robô durante os testes realizados.

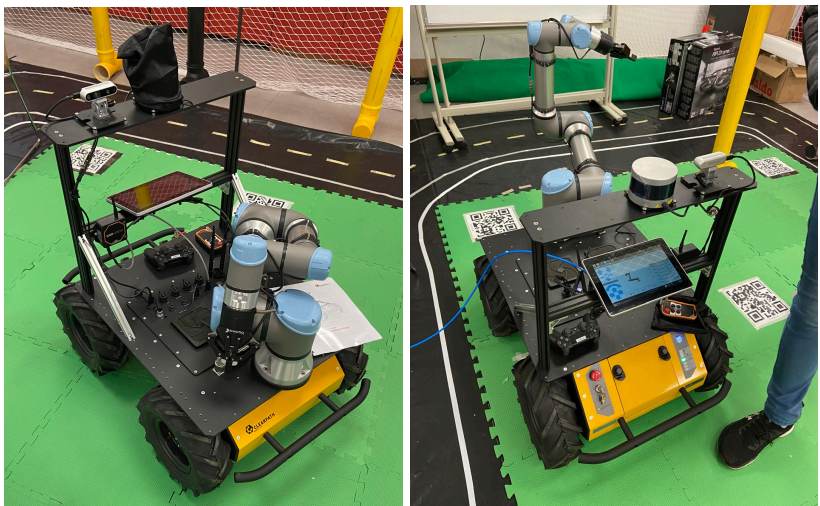

Figura 8. Testes de validação de funcionamento dos componentes.

### 3.3 - SIMULADOR DO ROBÔ

Além do suporte ao usuário para configuração do acesso ao robô, a empresa também disponibiliza um ambiente virtual de simulação desenvolvido para testes e pesquisa na área. Com ele é possível visualizar o robô e simular movimentos com obstáculos e variáveis físicas para desenvolver algoritmos de movimentação em diferentes ambientes. Foi desenvolvido um "Docker", um container que facilita a instalação do simulador, contendo as dependências necessárias para que o processo de simulação seja possível, incluindo as bibliotecas e requisitos. A continuação, na Figura 9 é possível visualizar o ambiente de simulação. (Los-UruBots-Del-Norte, s. f.)

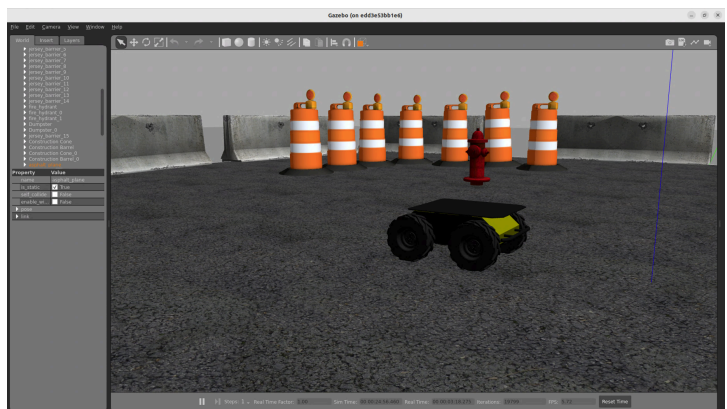

Figura 9. Ambiente de simulação Gazebo para robô Husky.

### 4 - RESULTADOS

Como resultado deste trabalho, a verificação de funcionamento dos componentes do robô adquirido foi realizada. Foram realizados testes de acesso ao robô, funcionamento dos sensores e atuadores, como leitura com o sensor *Lidar VLP-16*, câmera *Intel RealSense D435i*, movimentação com a plataforma móvel, manipulador e verificação de status de baterias e outros componentes. Na Figura 10 a continuação é possível visualizar um dos processos de funcionamento da pinça do manipulador e nas Figura 11, as validações de funcionamento de movimentação e status de componentes.

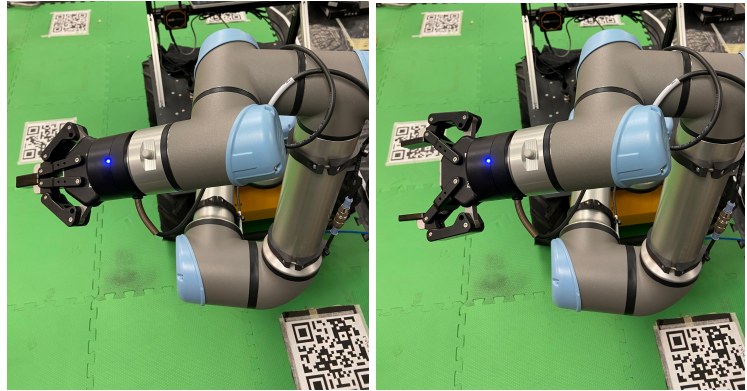

Figura 10. Resultados de funcionalidade da pinça.

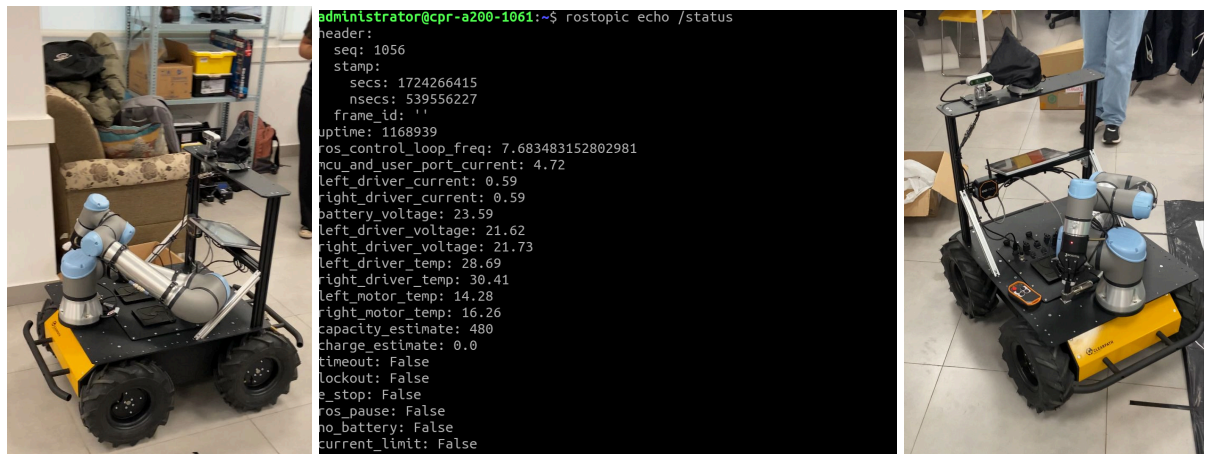

Figura 11. Resultados de movimentação e status de funcionamento.

## 5 - CONCLUSÕES

Este estudo detalha a validação técnica de operação da plataforma robótica Husky e suas extensões para um contexto de pesquisa e de processo de aquisição da Universidade Tecnológica do Uruguai. Com este trabalho foi possível explorar as funcionalidades do robô e compreender quais são as áreas potenciais de exploração da plataforma robótica e suas extensões. O robô husky em questão, conta com dispositivos com uma das mais avançadas tecnologias em questão de movimentação e detecção de objetos do mundo, permitindo a implementação de algoritmos com baixa porcentagem de erro de hardware ao comparar simulações e aplicações reais.

A validação realizada permite a exploração de diferentes áreas em quesito, trazendo aos autores envolvidos um conhecimento mais profundo da área técnica de seu trabalho desenvolvido no laboratório em questão. Isso permite que se capacite cada vez mais as pessoas do entorno com conhecimentos para contatos com empresas da indústria, que se desenvolve a cada ano com tecnologias de alto nível e alto valor agregado.

Pela parte de contribuição do trabalho, a realização da configuração inicial da plataforma robótica e seus componentes resultou em um guia em português para facilitação de novos pesquisadores no processo de verificação de utilidade do robô em questão por completo, tanto em ambiente simulado como em testes reais. Este guia permite ao usuário que localize possíveis erros entre etapas de utilização e detecção de características importantes do equipamento.

# 6- REFERENCIAS